\documentclass[journal]{IEEEtran}
\ifCLASSOPTIONcompsoc
\else
\fi
\ifCLASSINFOpdf
\else
\fi
\usepackage{graphicx}
\usepackage{amssymb}
\usepackage{amsmath}
\usepackage{bm}
\usepackage[Algorithm, boxed]{algorithm}
\usepackage{algorithmic}
\usepackage[bookmarks=false]{hyperref}
\usepackage{breakurl}
\usepackage{multirow}
\usepackage{CJK}
\begin{document}
%
\title{A study on cost behaviors of binary classification measures 
in class-imbalanced problems}
%
%
%
%

\author{Bao-Gang Hu,~\IEEEmembership{Senior Member,~IEEE},
~ Wei-Ming Dong, ~\IEEEmembership{Member,~IEEE}
\begin{CJK*}{GB}{gbsn}\author{Pinyin Name (???)}\end{CJK*}
\IEEEcompsocitemizethanks{\IEEEcompsocthanksitem B.-G. Hu and W.-M. Dong 
are with NLPR/LIAMA,
Institute of Automation, Chinese Academy of Sciences, Beijing 100190, China.\protect\\
E-mail: hubg@nlpr.ia.ac.cn \protect\\
E-mail: weiming.dong@ia.ac.cn
}
\thanks{}}

\IEEEcompsoctitleabstractindextext{%
\begin{abstract}
This work investigates into cost behaviors of binary 
classification measures in a background of class-imbalanced problems. 
Twelve performance measures are studied, such
as $F$ measure,  G-means in terms of accuracy rates, 
and of recall and precision, balance error rate ($BER$), 
Matthews correlation coefficient ($MCC$), Kappa coefficient ($\kappa$), etc.
A new perspective is presented 
for those measures by revealing their cost
functions with respect to the class imbalance ratio. 
Basically, they are described by four types of 
cost functions. The functions provides 
a theoretical understanding why some measures are 
suitable for dealing with class-imbalanced problems.
Based on their cost functions, we are able to conclude
that G-means of accuracy rates and $BER$ 
are suitable measures because they 
show {\it ``proper'' }  cost behaviors in terms of
{\it ``a misclassification from a small class 
will cause a greater cost than that from a large class''. } 
On the contrary, $F_1$ measure,  G-means of recall and precision,
$MCC$ and $\kappa$ measures do not produce such behaviors
so that they are unsuitable to serve our goal 
in dealing with the problems properly.

\end{abstract}

}

\maketitle

\IEEEdisplaynotcompsoctitleabstractindextext

%
\IEEEpeerreviewmaketitle

\section{Introduction}
 
Class-imbalanced problems become more common and serious 
in the emergence of {\it ``Big Data''} processing. The initial reason 
is due to a fact that useful information is generally 
represented by a minority class. Therefore, the 
{\it class-imbalance} (or {\it skewness}) {\it ratio} between a {\it majority} class over 
a {\it minority} 
one can be severely large \cite{Chawla}. The other reason can be 
appeared from utilizations of {\it ``one-versus-rest''} binary 
classification scheme for a fast processing of 
multiple classes \cite{Akata}. 
Generally, the greater the number of classes, the larger the 
class-imbalance ratio.
When most investigations in the conventional classifications 
apply {\it accuracy} (or {\it error}) {\it rate} as a learning criterion, this performance
measure is no more appropriate in 
dealing with highly-imbalanced datasets \cite{He}. 
In addressing class-imbalanced problems properly, {\it cost-sensitive 
learning} is proposed in which users are required to specify the 
costs according to error types \cite{Elkan}. At the same time, the other investigations 
apply {\it ``proper"} 
measures \cite {Weiss}, or learning criteria, which do not require  information about
costs. Those measures, such as $F$-measures, $AUC$ and $G$-means, are 
considered to be {\it cost-free learning} \cite{Zhang}. 
Significant progresses have been reported on using those measures
\cite {Kubat, Daskalaki, Huang, Menon}. 
Within the classification studies, however, we consider that two important issues
below are still unclear theoretically, that is:

 {\it I. Why some of measures are successful in dealing with highly-imbalanced datasets? 

II. What are the function behaviors 
of binary classification measures when the class-imbalance ratio increases? } 

The questions above form the motivation of this work. 
In principle, we can view that any classification measure implies 
cost information even one does not specify it explicitly.  
Taking a measure of error rate for example. When this
measure is set as a learning criterion in binary 
classifications, a {\it ``zero-one''} cost
function is given to the criterion \cite{Duda}. 
This function assigns an equal cost to both errors
from two classes. Therefore,  
a new perspective from the cost behaviors is proposed
in this study in order
to answer the questions.
Twelve measures are selected in this study on
binary classifications. 
The rest of this brief paper is organized as follows.
In Section II, we discuss two levels of evaluations in the selection of measures.  
Twelve measures in binary classifications are 
presented in Section III. Their cost functions
are derived in Section IV. We demonstrate
numerical examples in Section V. The 
conclusions are given in Section VI.

\section{Function-based vs performance-based evaluations}

This section will discuss measure selection
in classifications. Fig. 1 shows two levels of evaluations, namely,
{\it function-based} and {\it performance-based} evaluations.
From an application viewpoint, the performance-based evaluation
seems more common because it can provide a fast and overall picture
among the candidate measures. One of typical investigations
is shown by Ferri et al \cite{Ferri} on eighteen performance measures
over thirty datasets. However, this kind of investigations 
generally produce the performance responses, 
not only to the measures, but also to the data and 
associated learning algorithms. Therefore, conclusions
from the performance-based evaluation
may be changed accordingly with the different datasets.
Due to the coupling feature in the performance responses, one may fail to
obtain the intrinsic properties of the measures. 

We consider that the function-based evaluation
is more fundamental in the measure selection. 
This evaluation will reveal 
{\it function} (or {\it property}) {\it differences} among the measures.
Without involving any 
learning algorithm and noisy data, 
one is able to gain the intrinsic properties
of measures. The properties can 
be various depending on the specific concerns,
such as, ROC isometrics \cite{Flach},
statistical properties of AUC measure \cite{Ling}, 
monotonicity and error-type differentiability \cite{Leichter}.
According to a specific property, one is able to see why one 
measure is more {\it ``proper"} than the others. 
The findings from the function-based evaluation
will be independent of the learning algorithms and datasets. 

\begin{figure}[htb]
    \centering{
    \includegraphics[width=88mm]{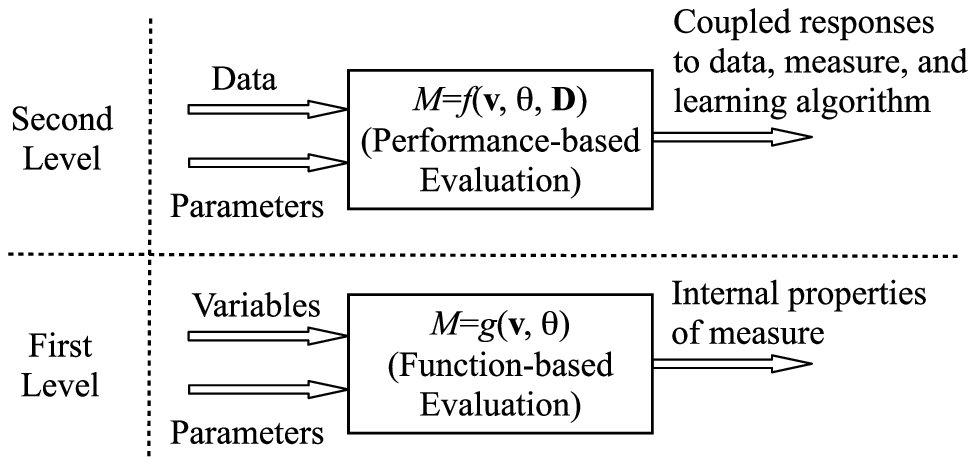} \newline \newline 
    \small Fig. 1. Schematic diagram of two levels of 
    evaluations in the measure selections. ~~~~~~~}
\end{figure}

In this work, we will focus on a specific property which
is not well studied in the function-based evaluation. 
Suppose that any binary classification measure 
produces cost functions in an implicit form. We consider 
a measure to be {\it ``proper''} for processing 
class-imbalanced problems only when it holds a
{\it ``desirable''} property so that {\it ``a misclassification from a small class 
will cause a greater cost than that from a large class''}  \cite{Hu}.
We call this property to be a 
{\it ``meta measure''} because it describes high-level
or qualitative knowledge about a specific measure. 
If a binary classification measure satisfies (or does not satisfy) the 
meta measure, we call it {\it ``proper''} (or {\it ``improper''}).
The examination in terms of the meta-measure enables clarification 
of the intrinsic causes of performance differences among classification measures.

\section{Two-class measures}

A binary classification is considered in this
work, and it is given by a {\it confusion matrix} $\textbf{C}$
in a form of:
\begin{equation}
\textbf{C} = \left[ {\begin{array}{*{20}c}
   {TN} & {FP }    \\
   {FN} & {TP }  \\
\end{array}} \right],
\end{equation}
where ``\emph{TN}", ``\emph{TP}", ``\emph{FN}", ``\emph{FP}",
represent ``\emph{true negative}" , ``\emph{true positive}",
``\emph{false negative}", ``\emph{false positive}", respectively.
Suppose $N$ ($= TN+TP+FN+FP$) to be the {\it total number} of samples in the
classification. The confusion matrix can be shown in the
other form:
\begin{equation}
\textbf{C} = N \left[ {\begin{array}{*{20}c}
   {CR_1} & {E_1 }    \\
   {E_2} & {CR_2}  \\
\end{array}} \right],
\end{equation}
where 
$CR_1$, $CR_2$, $E_1$, and $E_2$ are the {\it correct recognition rates} 
and {\it error rates} \cite {Hu} of Class 1 and Class 2,
respectively. They are defined by:
\begin{equation}
CR_1=  \frac{TN} {N}, ~CR_2=  \frac{TP} {N},
\end{equation}
\begin{equation}
E_1=  \frac{FP} {N} , ~E_2=  \frac{FN} {N} ,
\end{equation}
and form the relations to the {\it population rates} by:
\begin{equation}
p_1=CR_1+E_1, ~p_2=CR_2+E_2.
\end{equation}
From the non-negative terms in the confusion matrix, one can get the
following constraints:
\begin{equation}
 \begin{array}{r@{\quad}l}
& 0 < p_1 <1, ~~ 0 < p_2 <1, ~ p_1+p_2=1 \\
& 0 \leq E_1 \leq p_1, ~ 0 \leq E_2 \leq p_2. 
\end{array} 
\end{equation}

Twelve measures are investigated in this work. 
The first measure is the {\it total accuracy rate}:
\begin{equation}
A_T=  \frac{TN+TP} {N} = 1 -E_1-E_2.
\end{equation}
In this work, we will adopt the notions of four means (Fig. 2),
namely, {\it Arithmetic Mean}, {\it Geometric Mean}, {\it Quadratic Mean}
and {\it Harmonic Mean}, in constructions of performance measures.
 

\begin{figure}[htb]
    \centering{
    \includegraphics[width=62mm]{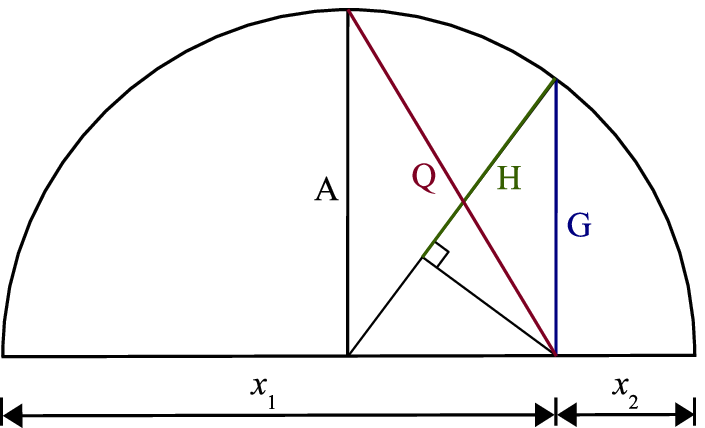} \newline \newline 
    \small Fig. 2. Graphical interpretations of four means. ~~~~~~~}
\end{figure}

From the definitions of {\it precision} ($P$) and {\it recall} ($R$):
\begin{equation}
P=  \frac{TP} {TP+FP}=\frac{CR_2} {CR_2+E_1},~ R=  \frac{CR_2} {p_2},
\end{equation}
one can obtain four precision-recall-based means:
\begin{equation}
A_{PR}=(P+R)/2.
\end{equation}
\begin{equation}
G_{PR}=  \sqrt{PR}.
\end{equation}
\begin{equation}
Q_{PR}=  \sqrt{\frac{P^2 + R ^2} {2} }.
\end{equation}
\begin{equation}
H_{PR}=F_1= 2 \frac{P*R} {P+R}.
\end{equation}
Eq. (12) shows that $F_1$ measure is the harmonic mean of precision
and recall. More definitions are given below
\begin{equation}
 \begin{array}{r@{\quad}l}
& A_1=TNR= Specificity  =  \frac{TN} {TN+FP}=  \frac{CR_1} {p_1}, \\
& A_2=TPR= Sensitivity =  \frac{TP} {TP+FN} =  \frac{CR_2} {p_2}=R,
\end{array} 
\end{equation} 
where the {\it accuracy rate of the first class} ($A_1$)
can also be called {\it true negative rate} ($TNR$) or {\it specificity};
the {\it accuracy rate of the second class} ($A_2$) called 
{\it true positive rate} ($TPR$), {\it sensitivity} or {\it recall}. 
In this work, we adopt the term of {\it accuracy rate of the $i$th class} ($A_i$)
because it is extendable if multiple-class problems are considered. 
The relation between the total accuracy rate and 
the accuracy rate of the $i$th class is 
\begin{equation}
A_T =p_1*A_1+p_2*A_2.
\end{equation}

Then, four accuracy-rate-based means are formed as:
\begin{equation}
A_{A_i}=AUC_b=(A_1+A_2)/2.
\end{equation}
\begin{equation}
G_{A_i}=  \sqrt{A_1 * A_2}.
\end{equation}
\begin{equation}
Q_{A_i}=  \sqrt{\frac{A_1^2 + A_2 ^2} {2} }.
\end{equation}
\begin{equation}
H_{A_i}= 2 \frac{A_1*A_2} {A_1+A_2}.
\end{equation}
In eq.  (15), $AUC_b$ is the {\it area under the curve} ($AUC$) for a single classification
point in the $ROC$ curve. $AUC_b$ is also called {\it balanced accuracy} \cite{Velez}.  
Three other measures are also received attentions.  
The {\it balance error rate} ($BER$) is given in a form of:
\begin{equation}
BER=  \frac{1} {2}( \frac{E_1} {p_1} + \frac{E_2} {p_2}).
\end{equation}
The {\it Matthews correlation coefficient} ($MCC$) is given by:
\begin{equation}
MCC=  \frac{TP*TN-FP*FN} {\sqrt{p_1 p_2 N^2(TN+FN)(TP+FP)}}.
\end{equation}
The {\it Kappa coefficient} ($ \kappa $) is given by:
\begin{equation}
\begin{array}{r@{\quad}l}
& \kappa =  \frac{Pr(a)-Pr(e)} {1-Pr(e)}, \\
& Pr(a)=  \frac{TN+TP} {N},  \\
& Pr(e)=  p_1 \frac{TN+FN} {N} +p_2\frac{TP+FP} {N} .
\end{array} 
\end{equation}
One needs to note that 
the first ten measures are given in a range of [0, 1],
and the last two measures, $MCC$ and $\kappa$, are within a range of [-1,1]. 
When the four precision-recall-based measures do not take 
the true negative rate into account, all other measures do.
Some measures above may be not well adopted in applications.
We investigate them for the reason of a comparative study.

\section{Cost functions of measures}

The risk of binary 
classifications can be described by \cite{Duda}:
 \begin{equation}
Risk = \lambda_{11}CR_1 + \lambda_{12}E_1 + 
\lambda_{22}CR_2 + \lambda_{21}E_2,
\end{equation}
where $\lambda_{ij}$ is a cost term for the true class of a pattern 
to be $i$, but be misclassified as $j$. In the cost sensitive learning,
the cost terms are generally assigned with constants \cite{Elkan}. 
However, we consider all costs in binary classifications
can be described in a function form of $\lambda_{ij} (\textbf{v})$, 
where \textbf{v} is a variable vector. The size of the vector will
be discussed later. We call $\lambda_{ij} (\textbf{v})$ 
{\it ``cost function''}, or {\it ``equivalent cost''} if it
is not given explicitly. 
In the derivation of cost functions of the given
measures, we make several assumptions below:

\begin{itemize}
\item[] $\cal {A}$1. The basic information to 
derive the cost functions is a confusion matrix 
in a binary classification problem without a reject option. 
\item[] $\cal {A}$2. The population rate of the second class 
$p_2$ corresponds to the minority class,
that is, $p_2<0.5$. Hence, $p_1$ corresponds to the majority class,
\item[] $\cal {A}$3. For simplifying analysis without losing generality, 
we assume  
 $\lambda_{11}=\lambda_{22}=0$. Therefore,  
only $\lambda_{12} (\textbf{v})$ and $\lambda_{21} (\textbf{v})$ 
are considered, but required to be non-negative ($\geq 0$) for $Risk \geq 0$.
\item[] $\cal {A}$4. When the exact cost function cannot be 
obtained, the Taylor approximation will 
be applied by keeping the linear terms,
and neglecting the remaining higher-order terms. 
The function is then denoted by $ \hat{\lambda}_{ij} (\textbf{v})$.
\end{itemize}

When all the measures, except $BER$, are given in a maximum sense to the task of
classifications, we need to transfer them into the minimum sense
in the form of eq. (22). This transformation should not destroy
the evaluation conclusions. For example, we can find an equivalent
relation between the total accuracy rate and error rates:
\begin{equation}
max ~ A_T ~ \Leftrightarrow ~ min ~ {\cal R}isk ~(A_T) = E_1 + E_2
\end{equation} 
where {\it ``max''} and
{\it ``min''} are denoted {\it ``maximization''} and {\it ``minimization''} operators,
respectively;
the symbol ``$\Leftrightarrow$'' is for {\it ``equivalency''}; and 
``${\cal R}isk$'' is the transformation operator. 
Using the expression of eq. (22), one can immediately obtain
the equivalent costs for the accuracy measure, 
$\lambda_{12}=\lambda_{21}=1 $. The costs indicate 
constant values and no distinction
between two types of errors. 

However, in most cases, one fails to obtain the exact expressions on
$\lambda_{ij}$. One example is given on the general form of $F$ measure 
by a transformation \cite{Martino}:
\begin{equation}
\begin{array}{r@{\quad}l}
& max ~ F_\beta = (1+\beta ^2 )\frac{PR} {\beta ^2 P+R} ~ \Leftrightarrow \\
& min ~ {\cal R}isk ~(F_\beta) = \frac{E_1} {p_2-E_2} + \frac{\beta ^2 E_2} {p_2-E_2} ,
\end{array} 
\end{equation} 
from which we can only get so called {\it ``apparent cost functions''} in a form of:
\begin{equation}
\lambda_{12}^A=\frac{1} {p_2-E_2}, ~ \lambda_{21}^A= \frac{\beta ^2} {p_2-E_2}.
\end{equation} 
The term of {\it ``apparent''} 
is used because the exact functions 
without coupling with $E_i$ may never be obtained from the given measure.
Hence, the apparent cost functions in binary classifications without a reject option
can be described in a general form of:
\begin{equation}
\lambda_{ij}^A = \lambda_{ij}^A (E_1, E_2,p_2). 
\end{equation}
From the relations of eqs. (2)-(6), only three
independent variables are used in describing the 
functions. One can apply the {\it ``class imbalance} (or {\it skewness}) 
{\it ratio"},
$S_r=p_1 / p_2$, to replace the variable $p_2$
for the analysis. 
The apparent cost functions provide users an 
analytical power in terms of a complete set of independent variables.

However, one is unable to realize unique representations
of costs, either exact or apparent, on all measures, 
such as on $G_{Ai}$ or $G_{PR}$.
For overcoming this difficulty, we adopt a strategy of
the first-order approximation, $\cal {A}$4.
Therefore, one will get a general form of $ \hat{\lambda}_{ij} (p_2) $
with only a single variable for binary classifications. 
From the relation \cite{Elkan} of $min ~ Risk \Leftrightarrow min ~ a*Risk + b$, 
the constants $a$ and $b$ will be removed in the
derivation of $ \hat{\lambda}_{ij} (p_2) $,
which will not destroy the classification
conclusions.
 
Table I lists the all measures and their cost functions or values. 
Only three measures exist the exact solutions on the costs. 
The other measures, originally given in a form of 
maximization sense in classifications, need to be
transformed into a minimization sense.
Suppose $M$ to be one of those measures, 
we adopt the following transformation:
\begin{equation}
{\cal R}isk ~(M) = \frac{1} {M-M_{min}},
\end{equation}
where $M_{min}$ is the minimum value of $M$.
The transformation above is meaningful on
three aspects. First, it keeps classification
conclusions invariant.
Second, it satisfies the assumption of $Risk \geq 0$ because $M-M_{min}\geq0$.
Third, it can describe an infinitive risk when $M=M_{min}$.

\begin{table*}[htbp]
\caption{twelve measures and their cost functions.}
\centering
\begin{tabular}{lllll}
\hline
Name of measures  & Calculation & Cost  & When & Remark on\\
  $[$Main reference$]$  & formulas & functions & $p_2 \rightarrow 0$ & cost functions\\
\hline
 \parbox[c][1.1cm]{3.0cm }{Total accuracy rate \cite{Duda}} 
& $ A_T=  1-E_1-E_2 $ 
& \parbox[c][1.1cm]{2.7cm }{$ \lambda_{12}=1 $ \\ $ \lambda_{21}=1 $}   
& \parbox[c][1.1cm]{1.8cm }{$ \lambda_{12}=1 $ \\ $ \lambda_{21}=1 $}   
& \parbox[c][1.1cm]{2.4cm }{Exact \\ costs} \\
\hline
\parbox[c][1.1cm]{3.0cm }{Arithmetic mean of \\  precision and recall \cite{Henderson}}   
& $A_{PR}=\frac{P+R} {2} $ 
& \parbox[c][1.1cm]{2.7cm }{$ \hat{\lambda}_{12}=\frac{1} {p_2} $ \\ $ \hat{\lambda}_{21}=\frac{1} {p_2} $}  
& \parbox[c][1.1cm]{1.8cm }{$ \hat{\lambda}_{12} \rightarrow \infty $ \\ $ \hat{\lambda}_{21} \rightarrow \infty $}   
& \parbox[c][1.1cm]{2.4cm }{Lower bounds  \\ if $E_1 > (2+\sqrt{5})E_2$ } \\
\hline
\parbox[c][1.1cm]{3.0cm }{Geometric mean of \\  precision and recall \cite{Daskalaki}}  
& $G_{PR}=  \sqrt{P R}$  
& \parbox[c][1.1cm]{2.7cm }{$ \hat{\lambda}_{12}=\frac{1} {p_2} $ \\ $ \hat{\lambda}_{21}=\frac{1} {p_2} $}    
& \parbox[c][1.1cm]{1.8cm }{$ \hat{\lambda}_{12} \rightarrow \infty $ \\ $ \hat{\lambda}_{21} \rightarrow \infty $}   
& \parbox[c][1.1cm]{3.1cm }{Lower bounds  \\ if $E_1 > (3+2\sqrt{3})E_2$ } \\
\hline
\parbox[c][1.1cm]{3.0cm }{Quadratic mean of \\ precision and recall \cite{Kan}}  
& $Q_{PR}=  \sqrt{\frac{P^2 + R ^2} {2} }$ 
& \parbox[c][1.1cm]{2.7cm }{$ \hat{\lambda}_{12}=\frac{1} {p_2} $ \\ $ \hat{\lambda}_{21}=\frac{1} {p_2} $}  
& \parbox[c][1.1cm]{1.8cm }{$ \hat{\lambda}_{12} \rightarrow \infty $ \\ $ \hat{\lambda}_{21} \rightarrow \infty $}   
& \parbox[c][1.1cm]{3.1cm }{Lower bounds  \\ if $E_1 > (\frac{5} {3}+\frac{2} {3}\sqrt{7})E_2$ } \\
\hline
\parbox[c][1.2cm]{3.0cm }{Harmonic mean of \\  precision and recall \\ (or $F_1$ measure) \cite{Rijsbergen} }  
& $H_{PR}=F_1 = 2 \frac{P*R} {P+R}$ 
& \parbox[c][1.1cm]{2.7cm }{$ \hat{\lambda}_{12}=\frac{1} {p_2} $ \\ $ \hat{\lambda}_{21}=\frac{1} {p_2} $}  
& \parbox[c][1.1cm]{1.8cm }{$ \hat{\lambda}_{12} \rightarrow \infty $ \\ $ \hat{\lambda}_{21} \rightarrow \infty $}   
& \parbox[c][1.1cm]{2.4cm }{Lower bounds \\ for any $E_i$} \\
\hline
 \parbox[c][1.1cm]{3.0cm }{Arithmetic mean of \\  accuracy rates \cite{Velez}} 
& $ A_{A_i}=AUC_b=(A_1+A_2)/2 $ 
& \parbox[c][1.1cm]{2.7cm }{$ \lambda_{12}=\frac{1} {1-p_2} $ \\ $ \lambda_{21}=\frac{1} {p_2} $}   
& \parbox[c][1.1cm]{1.8cm }{$ \lambda_{12}=1 $ \\ $ \lambda_{21} \rightarrow \infty $}   
& \parbox[c][1.1cm]{2.4cm }{Exact \\ functions} \\
\hline
\parbox[c][1.1cm]{3.0cm }{Geometric mean of \\  accuracy rates \cite{Kubat}}  
& $G_{A_i}=  \sqrt{A_1 * A_2}$ 
& \parbox[c][1.1cm]{2.7cm }{$ \hat{\lambda}_{12}=\frac{1} {1-p_2} $ \\ $ \hat{\lambda}_{21}=\frac{1} {p_2} $}  
& \parbox[c][1.1cm]{1.8cm }{$ \hat{\lambda}_{12} =1 $ \\ $ \hat{\lambda}_{21} \rightarrow \infty $}   
& \parbox[c][1.1cm]{2.4cm }{Lower bounds \\ for any $E_i$} \\
\hline
\parbox[c][1.1cm]{3.0cm }{Quadratic mean of \\  accuracy rates \cite{Liu}}  
& $ Q_{A_i}=  \sqrt{\frac{A_1^2 + A_2 ^2} {2}}$ 
& \parbox[c][1.1cm]{2.7cm }{$ \hat{\lambda}_{12}=\frac{1} {1-p_2} $ \\ $ \hat{\lambda}_{21}=\frac{1} {p_2} $}   
& \parbox[c][1.1cm]{1.8cm }{$ \hat{\lambda}_{12} =1 $ \\ $ \hat{\lambda}_{21} \rightarrow \infty $}   
& \parbox[c][1.1cm]{2.4cm }{Lower bounds \\ for any $E_i$} \\
\hline
\parbox[c][1.1cm]{3.0cm }{Harmonic mean of \\ accuracy rates \cite{Kennedy}}  
& $H_{A_i}= 2 \frac{A_1*A_2} {A_1+A_2}$ 
& \parbox[c][1.1cm]{2.7cm }{$ \hat{\lambda}_{12}=\frac{1} {1-p_2} $ \\ $ \hat{\lambda}_{21}=\frac{1} {p_2} $}   
& \parbox[c][1.1cm]{1.8cm }{$ \hat{\lambda}_{12} =1 $ \\ $ \hat{\lambda}_{21} \rightarrow \infty $}   
& \parbox[c][1.1cm]{2.4cm }{Lower bounds \\ for any $E_i$} \\
\hline
\parbox[c][1.1cm]{3.0cm }{Balance error  \\ rate (BER) \cite{Guyon}}   
& $BER=  \frac{1} {2}( \frac{E_1} {p_1} + \frac{E_2} {p_2})$ 
& \parbox[c][1.1cm]{2.7cm }{$ \lambda_{12}=\frac{1} {1-p_2}$ \\ $ \lambda_{21}=\frac{1} {p_2} $}    
& \parbox[c][1.1cm]{1.8cm }{$ \lambda_{12} =1 $ \\ $ \lambda_{21} \rightarrow \infty $}   
& \parbox[c][1.1cm]{2.4cm }{Exact \\ functions} \\
\hline
\parbox[c][1.1cm]{3.0cm }{Matthews correlation \\ coefficient (MCC) \cite{Baldi}}    
& $MCC=  \frac{TP*TN-FP*FN} {\sqrt{p_1 p_2 N^2(TN+FN)(TP+FP)}}$ 
& \parbox[c][1.1cm]{2.7cm }{$ \hat{\lambda}_{12}=\frac{1} {p_2(1-p_2)} $ \\ $ \hat{\lambda}_{21}=\frac{1} {p_2(1-p_2)} $}  
& \parbox[c][1.1cm]{1.8cm }{$ \hat{\lambda}_{12} \rightarrow \infty $ \\ $ \hat{\lambda}_{21} \rightarrow \infty $}   
& \parbox[c][1.1cm]{3.1cm }{Unknown for \\  bound features } \\
\hline
\parbox[c][1.1cm]{3.0cm }{Kappa coefficient ($\kappa$) \\  \cite{Cohen}}    
& $\kappa=  \frac{TN+TP-p_1(TN+FN)-p_2(TP+FP)} {N-p_1(TN+FN)-p_2(TP+FP)}$ 
& \parbox[c][1.1cm]{2.7cm }{$ \hat{\lambda}_{12}=\frac{1} {p_2(1-p_2)} $ \\ $ \hat{\lambda}_{21}=\frac{1} {p_2(1-p_2)} $}  
& \parbox[c][1.1cm]{1.8cm }{$ \hat{\lambda}_{12} \rightarrow \infty $ \\ $ \hat{\lambda}_{21} \rightarrow \infty $}   
& \parbox[c][1.1cm]{3.1cm }{Unknown for \\  bound features } \\
\hline
\end{tabular}
\end{table*}

From Table I, one can observe that the all measures investigated
in this work can be classified by 
four types of cost functions. Fig. 3 depicts 
the functions with respect to a single independent variable $p_2$.
We will discuss the cost behaviors according to
the function types first, and then the specific measures. 

\begin{figure}[htb]
    \centering{
    \includegraphics[width=82mm]{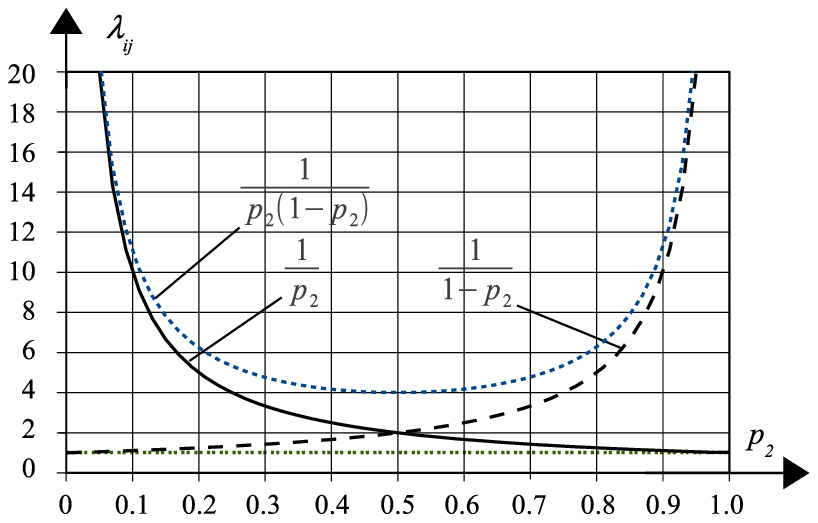} \newline
    \small Fig. 3. Plots of cost functions with respect to $p_2$. \newline
(Black-Solid:  $\lambda_{ij}  = \frac{1} {p_2} $. 
 Black-Dash: $\lambda_{ij}  = \frac{1} {1-p_2} $. 
 Blue-Dot: $\lambda_{ij}  = \frac{1} {p_2(1-p_2)} $.
 Green-Dot: $\lambda_{ij}  = 1 $.)
}
\end{figure}

Type I: $\lambda_{12}=\lambda_{21}=\lambda > 0 $. 

The costs are positive constants with equality. 
The classification solutions will be independent
of the constant values of costs whenever their equality relation holds. 
According to the meta measure,
this feature suggests that the total accuracy (or error) rate measure be 
{\it ``improper"} for dealing with
class-imbalanced problems. 

Type II: $ {\lambda}_{12}={\lambda}_{21}=\frac{1} {p_2} $.

Within this type of cost functions, both types of errors show the same cost behaviors
with respect to the $p_2$. It indicates no distinctions 
between two types of errors, which can be considered as
an {\it ``improper"} feature in class-imbalanced problems.   
Four measures from the precision-recall-based means 
demonstrate the same approximation expressions of $ \hat{\lambda}_{12}=
\hat{\lambda}_{21}=\frac{1} {p_2} $ as the lower bounds to the 
exact functions (Table I). 
However, their approximation rates are different and are not
given for the reason of their tedious expressions. 
The feature of the lower bounds will support the conclusions about
the cost behaviors of their exact functions on:
$ {\lambda}_{12}$ and ${\lambda}_{21}  \rightarrow \infty$ when $ {p_2}  \rightarrow 0$.
Another important feature is that this type of functions 
is {\it asymmetric} and
imposes more costs
on the positive class than on the
negative class. For example, from eq. (25), $F_1$ measure shows smaller costs of
$ {\lambda}_{12}$ = ${\lambda}_{21} =\frac{1} {1-E_2}$
if $ {p_1} = 0$.

Type III: $ {\lambda}_{12}=\frac{1} {1-p_2} , ~{\lambda}_{21}=\frac{1} {p_2} $.

This type of cost functions shows a {\it ``proper''} feature in processing 
class-imbalanced problems, because it satisfies the meta measure.
One can observe that in Fig. 3, when $p_2$ decreases, Type II error will receive
a higher cost than Type I error. 
Only when two classes are equal (also called {\it ``balanced''}), two types of errors 
will share the same values of costs. 
Note that the meta measure implies such requirement. 
Four measures from the accuracy-rate-based means and $BER$ measure
are within this type of the functions. 
In a study of the cost-sensitive learning, this type of the functions
can be viewed a {\it ``rebalance''} approach \cite{Elkan,Weiss,Akata}.
The exact solutions of the cost
functions inform that $BER$ and $A_{A_i}$ ($=AUC_{b}$) are fully equivalent in 
classifications. Their equivalency can also be gained from a
relation of $BER=1-A_{A_i}$.
The other three measures,  $G_{A_i}$, $Q_{A_i}$ and
$H_{A_i}$, present only approximations to the exact cost functions. 
Their lower bound features guarantee the cost behaviors of their exact functions on
$ {\lambda}_{12}=1$ and ${\lambda}_{21}  \rightarrow \infty$ when $ {p_2}  \rightarrow 0$.
This type of functions shows {\it symmetric} cost behaviors for any class to be
a minority.
 
Type IV: $ {\lambda}_{12}= {\lambda}_{21}=\frac{1} {p_2(1-p_2)} $.

Both $MCC$ and $\kappa$ measures approximate this type of cost functions. 
Because the same functions are given for the two types of errors,
any measure within this category will be {\it ``improper"} for
processing class-imbalanced problems. The functions are {\it symmetric} to 
either class being a minority.

From the context of class-imbalanced problems, one can further
aggregate the four types of cost functions within two categories,
namely, {\it ``proper cost type"}  and {\it ``improper cost type"}.
We consider only Type III cost function falls in the proper cost type,
and all others belong to the improper cost type. 
Hence, one can reach 
the most important finding from the category discussions about each measure. 
For example, 
when the two geometric mean measures,  $G_{A_i}$
and $G_{PR}$, are applied in the class-imbalanced problems \cite{Kubat,
Daskalaki}, respectively, their intrinsic
differences are not well disclosed. The present cost function study reveals
their property differences about the cost response to the skewness ratio. 
When  $G_{A_i}$ satisfies the desirable feature on the costs, 
$G_{PR}$ does not hold such feature. To our best knowledge, this 
theoretical finding has not been reported before.

Further finding is gained on $F$ measure. 
This measure is initially proposed in the area of information retrieval \cite{Rijsbergen}
for an overall balance between precision and recall.
Recently, $F$ measure is adopted increasingly in the
study of class-imbalanced learning \cite{Dembczynski,Ye,Maratea}. 
When $F$ measure is designed by concerning a {\it positive} ({\it minority}) class 
correctly without taking the {\it negative} ({\it majority}) class into account
directly, it does not mean suitability in processing
highly-imbalanced problems. The cost function analysis above
confirms that $F$ measure is {\it ``improper"} in either
class to be a {\it minority} when its population approximates zero. 

\section{Numerical examples}

For a better understanding of the investigated measures, 
we present numerical examples within two specific scenarios below. 

{\it Scenario I: Class populations are given.}

Within this scenario,
only two measures, $BER$ and $F_1$,  are considered in the
investigation for the following reasons. 
First, we need to demonstrate the exact cost functions graphically.
When $BER$ is qualified to this aspect, 
$F_1$ can also present the exact cost values when $E_2$ is known in eq. (25). 
Second, $BER$ and $F_1$ measures are representative to 
be {\it ``proper cost type"} and {\it ``improper cost type"}
respectively in cost functions. They form the
 {\it baselines} for understanding the other measures.

In the numerical examples, we assume the following data:
\begin{equation}
\begin{split}
& N=10000,   E_1=0.1, E_2=\frac{p_2} {2} , \\
& p_2=[0.5,0.1,0.05,0.01,0.005,0.001],
\end{split}
\end{equation}
where $p_2$ is given in a vector form to present classification changes,
such as   
from the {\it ``balanced''} to the {\it ``minority''} and {\it ``rare''}
stages, respectively.

\begin{table*}[htbp]
\caption{Solution data of ``$\lambda_{ij}$ vs. $p_2$'' for $BER$ and $F_1$ measures
 from the given data in eq. (28). }
\label{tab:comdist} \centering \setlength{\tabcolsep}{1.345pc}
\begin{tabular}{lcccccc}
\hline
 ~~ $p_2$  & 0.500 & 0.100 & 0.050 & 0.010 & 0.005 & 0.001\\
\hline
$BER$ & 0.350 & 0.306 & 0.303 & 0.301 & 0.300 & 0.300\\
\cline{2-7}
 $~~\lambda_{12}$ & 2.000 & 1.111 & 1.053 & 1.010 & 1.005 & 1.001\\
 \cline{2-7}
 $~~\lambda_{21}$ & 2.0 & 10.0 & 20.0 & 100.0& 200.0 & 1000.0\\
\hline
$F_1$ & 0.588 & 0.400 & 0.286 & 0.087 & 0.047 & 0.010\\
\cline{2-7}
 $~~\lambda_{12}$ & 4.0 & 20.0 & 40.0 & 200.0 & 400.0 & 2000.0\\
 \cline{2-7}
 $~~\lambda_{21}$ & 4.0 & 20.0 & 40.0 & 200.0 & 400.0 & 2000.0\\
\hline
\end{tabular}
\end{table*}

Table II shows the solutions to the given data in (28) for both
$BER$ and $F_1$ measures. The data of $BER$ and $F_1$ are 
calculated directly from the equations defined. 
The data of $\lambda_{ij}$ are the exact values  to each measure, respectively. 
One is able to  confirm the correctness of $\lambda_{ij}$ data through
the following relations:
\begin{equation}
BER=\frac{1} {2}(\lambda_{12}*E_1+ \lambda_{21}*E_2).
\end{equation}
\begin{equation}
\frac{1} {F_1} =1+\frac{1} {2}(\lambda_{12}*E_1+ \lambda_{21}*E_2).
\end{equation}

From the data in Table II, we can depict the 
plots of ``$\lambda_{ij}$ vs. $p_2$'' for $BER$ and $F_1$ measures (Fig. 4).
One can observe that $F_1$ measure is unable to distinct the costs,
but produces the same costs 
on the given data when $p_2$ decreases. 
Although $F_\beta$ can generate different
cost functions shown in (25) when $\beta \ne 1$,
the infinity feature still remains  
in the both cost functions if $p_2=0$.
This numerical example is sufficient to conclude that
$F_1$, or the other measures having the similar feature, is not suitable for processing 
class-imbalanced problems. 
On the contrary,
the cost plots of $BER$ measure confirm
the theoretical findings in the previous section. 
Among the twelve measures investigated, 
the measures within Type III cost functions
will exhibit the {\it ``proper"} cost behaviors in compatible
with our intuitions for solving class-imbalanced problems.

\begin{figure}[htb]
    \centering{
    \includegraphics[width=82mm]{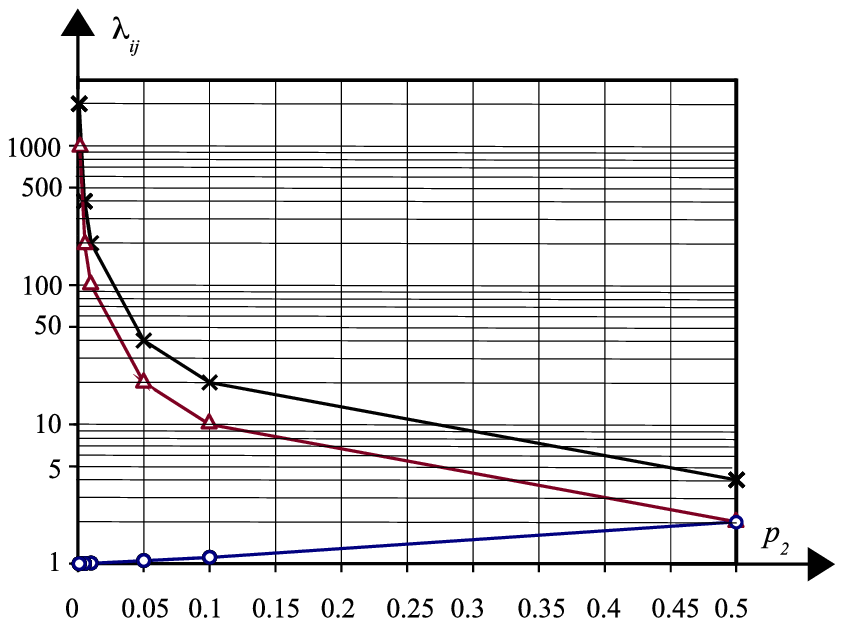} \newline
    \small Fig. 4. Plots of ``$\lambda_{ij}$ vs. $p_2$'' for $BER$ and $F_1$ measures
    on the given data in eq. (28). \newline
(Black-Cross:  $\lambda_{12}  = \lambda_{21}  = \frac{1} {p_2-E_2} $ 
for $F_1$ measure. \newline
Red-Triangle: $\lambda_{21}  = \frac{1} {p_2} $,
 Blue-Circle: $\lambda_{12}  = \frac{1} {1-p_2} $ for $BER$ measure.
 )
}
\end{figure}

{\it Scenario II: Gaussian distributions are given.}

This scenario is designed for a class-imbalance
learning. A specific set of Gaussian distributions is
exactly known, 
\begin{equation}
\begin{split}
& \mu_{1}=-1, \mu_{2}=1,\sigma_{1}=\sigma_{2}=1, \\
& p_2=[0.5,0.1,0.01,0.001,0.0001, 0.00001],
\end{split}
\end{equation}
where $ \mu_{i}$ and $\sigma_{i}$
are the mean and standard deviation to the {\it i}th class.
Five measures, $A_T$, $BER$, $F_1$, $G_{Ai}$ and $G_{PR}$,  are considered
for a comparative study.  
Table III shows the {\it optimum} solutions to the given data in (31) from using
the five measures, respectively.
Based on the data in Table III, Fig. 5 depicts the plots of ``$\frac  {E_2} {p_2} $ vs. $\frac {p_1}  {p_2}$'' 
for the measures.
When the class-imbalance ratio $\frac {p_1}  {p_2}$ increases, the minority
class (or Class 2) is mostly misclassified for measures $A_T$, $F_1$ and $G_{PR}$.
The value of  $\frac  {E_2} {p_2} =1.0$ suggests a {\it complete misclassification} on
all samples in Class 2. 
In comparison, $BER$ and $G_{Ai}$ measures show a small constant value of $\frac  {E_2} {p_2} $ ($=0.1587$),
which implies a good protection on the minority class. The two measures share the same 
solutions for the given distribution data in eq. (31). 
One can show that, when $\sigma_{1} \neq \sigma_{2}$, $BER$ and $G_{Ai}$
will present the different constant values. It can be further proved that all measures in Type III
will produce a constant behavior shown in Fig. 5, because their decision boundaries, $x_{b}$, will be
independent with the population variables. 

\begin{table*}[htbp]
\caption{Optimum solutions using the five measures respectively
 to the given data in eq. (31). 
  \newline (The subscripts "max" and "min" 
  stand for maximum and minimum respectively.
  $x_{b}$ is a decision boundary.)}
\label{tab:comdist} \centering \setlength{\tabcolsep}{1.345pc}
\begin{tabular}{lcccccc}
\hline
 ~~ $p_2$  & 0.50000 & 0.10000 & 0.01000 & 0.00100 & 0.00010  & 0.00001\\
\hline
$(A_T)_{max}$ & 0.8413 & 0.9299 & 0.9905 & 0.9990 & 0.9999  & 0.9999\\
\cline{2-7}
 $~~x_{b}$ & 0.0 & 1.0986 & 2.2976 & 3.4534 & 4.6051  & 5.7564 \\
\cline{2-7}
 $E_{1}/p_{1}$ & 1.587e-1 & 1.792e-2 & 4.876e-4 & 4.226e-6 & 1.041e-8  & 7.070e-12\\
 \cline{2-7}
 $E_{2}/p_{2}$ & 0.1587 & 0.5393 & 0.9028 & 0.9929 & 0.9998  & 0.9999\\
\hline
$(BER)_{min}$ & 0.1587 & 0.1587 & 0.1587 & 0.1587 & 0.1587  & 0.1587\\
\cline{2-7}
 $~~x_{b}$ & 0.0 & 0.0 & 0.0 & 0.0 & 0.0  & 0.0 \\
\cline{2-7}
 $E_{1}/p_{1}$ & 0.1587 & 0.1587 & 0.1587 & 0.1587 & 0.1587  & 0.1587\\
 \cline{2-7}
 $E_{2}/p_{2}$ & 0.1587 & 0.1587 & 0.1587 & 0.1587 & 0.1587  & 0.1587\\
\hline
$(F_1)_{max}$ & 0.8443 & 0.6121 & 0.3211 & 0.1291 & 0.0420  & 0.0118\\
\cline{2-7}
 $~~x_{b}$ & -.1570 & 0.6893 & 1.4705 & 2.1167 & 2.6843  & 3.1948\\
\cline{2-7}
 $E_{1}/p_{1}$ & 1.996e-1 & 4.557e-2 & 6.746e-3 & 9.145e-4 & 1.147e-4 & 1.365e-5\\
 \cline{2-7}
 $E_{2}/p_{2}$ & 0.1236 & 0.3780 & 0.6810 & 0.8679 & 0.9539  & 0.9859\\
\hline
$(G_{Ai})_{max}$ & 0.8413 & 0.8413 & 0.8413 & 0.8413 & 0.8413  & 0.8413\\
\cline{2-7}
 $~~x_{b}$ & 0.0 & 0.0 & 0.0 & 0.0 & 0.0  & 0.0\\
\cline{2-7}
  $E_{1}/p_{1}$ & 0.1587 & 0.1587 & 0.1587 & 0.1587 & 0.1587  & 0.1587\\
 \cline{2-7}
 $E_{2}/p_{2}$ & 0.1587 & 0.1587 & 0.1587 & 0.1587 & 0.1587  & 0.1587\\
\hline
$(G_{PR})_{max}$ & 0.8450 & 0.6123 & 0.3211 & 0.1293 & 0.0436 & 0.0139\\
\cline{2-7}
$~~x_{b}$ & -.1946 & 0.6697 & 1.4826 & 2.0481 & 2.2840  & 2.3260\\
\cline{2-7}
 $E_{1}/p_{1}$ & 2.103e-1 & 4.749e-2 & 6.519e-3 & 1.151e-3 & 5.116e-4 & 4.407e-5  \\
  \cline{2-7}
 $E_{2}/p_{2}$ & 0.1161 & 0.3706 & 0.6853 & 0.8527 & 0.9004  & 0.9076\\
\hline
\end{tabular}
\end{table*}

\begin{figure}[htb]
    \centering{
    \includegraphics[width=82mm]{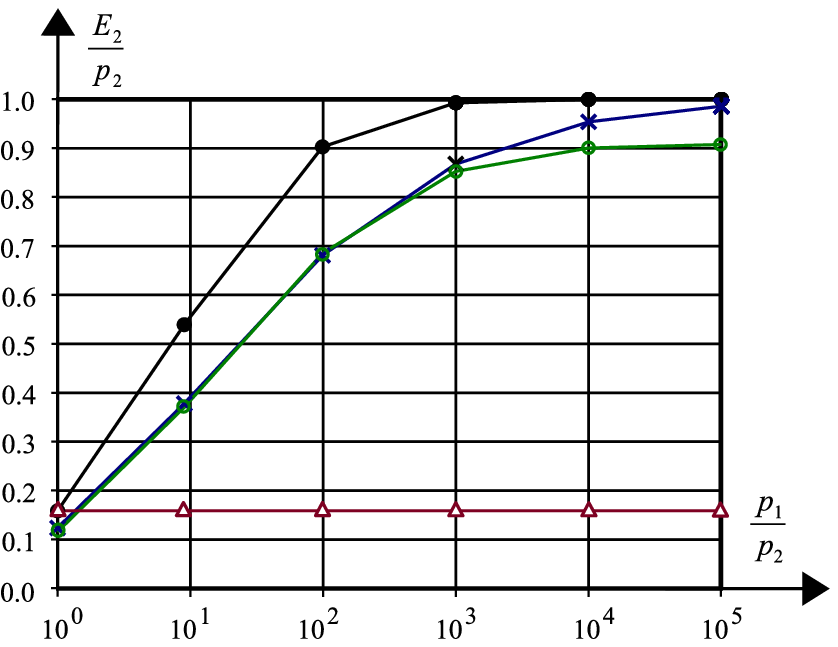} \newline
    \small Fig. 5. Plots of ``$\frac  {E_2} {p_2} $ vs. $\frac {p_1}  {p_2}$'' for five measures
    in Table III. \newline
(Black-Dot:  from $A_T$ measure.  Blue-Cross: from $F_1$ measure. \newline 
Green-Circle: from $G_{PR}$ measure.
Red-Triangle: from $BER$ and $G_{Ai}$ measures.
 )
}
\end{figure}

The numerical study in this scenario provides a counterexample to
confirm a general conclusion that $A_T$, $F_1$ and $G_{PR}$ are {\it ``improper"}
measures. If {\it ``improper"} measures are set as {\it ``learning targets}" (or {\it ``criteria}")
in highly-imbalanced problems, one may have a deleterious impact on
classification qualities. The numerical solutions of $BER$ and $G_{Ai}$ 
support the measures to be {\it ``proper"} only for the given
datasets. However, one is unable to reach a general conclusion on the two measures 
via numerical studies. 
This scenario study is also a function-based evaluation. If using real datasets for a performance-based evaluation, 
inconsistency findings may be introduced by population changes from sampling.  

\section{Conclusions}
\label{sec: con}

This work aims at developing a theoretical insight into why
some performance measures are appropriate, and some are not, for
solving class-imbalanced problems. Before reviewing the existing 
approaches, we discuss the
two levels of measure evaluations, that is, function-based
evaluation and performance-based evaluation. For revealing the
intrinsic properties of the measures, we consider the
function-based evaluation to be necessary, and investigate one
important aspect which is not well studied. This aspect
is defined to be the cost behaviors of binary classification 
measures in terms of class-imbalance skewness ratio. 
We adopt a {\it meta measure} in \cite{Hu} to examine each
measure to be {\it ``proper"} or {\it ``improper"} in 
applications.

Twelve measures are studied and their cost functions,
either exact or approximate, are derived. 
When four types of the cost functions are formed from the given measures, they
are basically two kinds according to the meta measure. 
The {\it ``proper"} kind includes 
the four means on accuracy rates and $BER$ (equivalently including $AUC_b$). 
The other measures,
i.e. $A_T$, the four means on precision and recall 
(including $F_1$), $MCC$ and $\kappa$, belong to 
{\it ``improper"} kind. 
Through the cost function analysis, one can observe their intrinsic
equivalences or differences among the measures.

In apart from the measures investigated in this work,
one can add other performance or meta measures for a systematic study. 
From an application viewpoint, we understand that
a final selection of measures (or learning criteria) may need to be based on 
an overall consideration regarding to each aspect in
function-based evaluation and performance-based
evaluation. 
The main point raised in this work confirms that
{\it ``what to learn (or learning-target selection)"} is the most imperative and primary issue 
in the study of machine learning.  
 
 \section*{Acknowledgment}

This work is supported in part by NSFC (No. 61273196)
for B.-G. Hu, and NSFC (No. 61172104) for W.-M. Dong.

%

\bibliographystyle{IEEETrans}


%





\end{document}